\documentclass{article}
\usepackage{spconf,amsmath,graphicx}
\usepackage{times}
\usepackage{epsfig}
\usepackage{graphicx}
\usepackage{amsmath}
\usepackage{amssymb}
\usepackage{xcolor}         
\usepackage{float} 
\usepackage{array,multirow}
\usepackage[accsupp]{axessibility}

\usepackage{booktabs}
\usepackage{tabularx}
\usepackage[breaklinks=true,bookmarks=false]{hyperref}
\usepackage{algorithm}
\usepackage[noend]{algpseudocode}

\title{Online Adaptive Personalization for Face Anti-spoofing}
\name{Davide Belli \quad Debasmit Das \quad Bence Major \quad Fatih Porikli}
\address{Qualcomm AI Research*\thanks{*Qualcomm AI Research is an initiative of Qualcomm Technologies, Inc.}\\
\tt\small \{dbelli, debadas, bence, fporikli\}@qti.qualcomm.com}
\begin{document}
\ninept
\maketitle
\begin{abstract}

Face authentication systems require a robust anti-spoofing module as they can be deceived by fabricating spoof images of authorized users. Most recent face anti-spoofing methods rely on optimized architectures and training objectives to alleviate the distribution shift between train and test users. However, in real online scenarios, past data from a user contains valuable information that could be used to alleviate the distribution shift. We thus introduce \textit{OAP} (\textit{Online Adaptive Personalization}): a lightweight solution which can adapt the model online using unlabeled data. OAP can be applied on top of most anti-spoofing methods without the need to store original biometric images. Through experimental evaluation on the SiW dataset, we show that OAP improves recognition performance of existing methods on both single video setting and continual setting, where spoof videos are interleaved with live ones to simulate spoofing attacks. We also conduct ablation studies to confirm the design choices for our solution.
\end{abstract}

\section{Introduction}
\vspace{-1mm}
\label{sec:intro}
Face authentication systems are widespread in everyday technology, but they can be easily spoofed by attackers when they have access to face images from the targeted user. Hence, face anti-spoofing models are an integral part of most modern face recognition systems. Face anti-spoofing research has received increased attention in recent years due to the availability of large-scale face image data, improvements in deep learning methods, and the potential for catastrophic data breaches. 
Nowadays, convolution neural networks~\cite{feng2016integration,li2016original,feng2020learning,yu2020searching} are a standard backbone for face anti-spoofing models, with recent work exploring additional modalities~\cite{kim2019basn,liu2018learning,kim2019basn,liu2018remote} or different task formulations like disentanglement~\cite{jourabloo2018face} and temporal modelling~\cite{yang2019face}.
However, most existing methods do not explicitly account for the distribution shift between training and testing face images.

 \begin{figure}[htb]
 \begin{minipage}[b]{1.0\linewidth}
   \centering
   \centerline{\includegraphics[width=\textwidth]{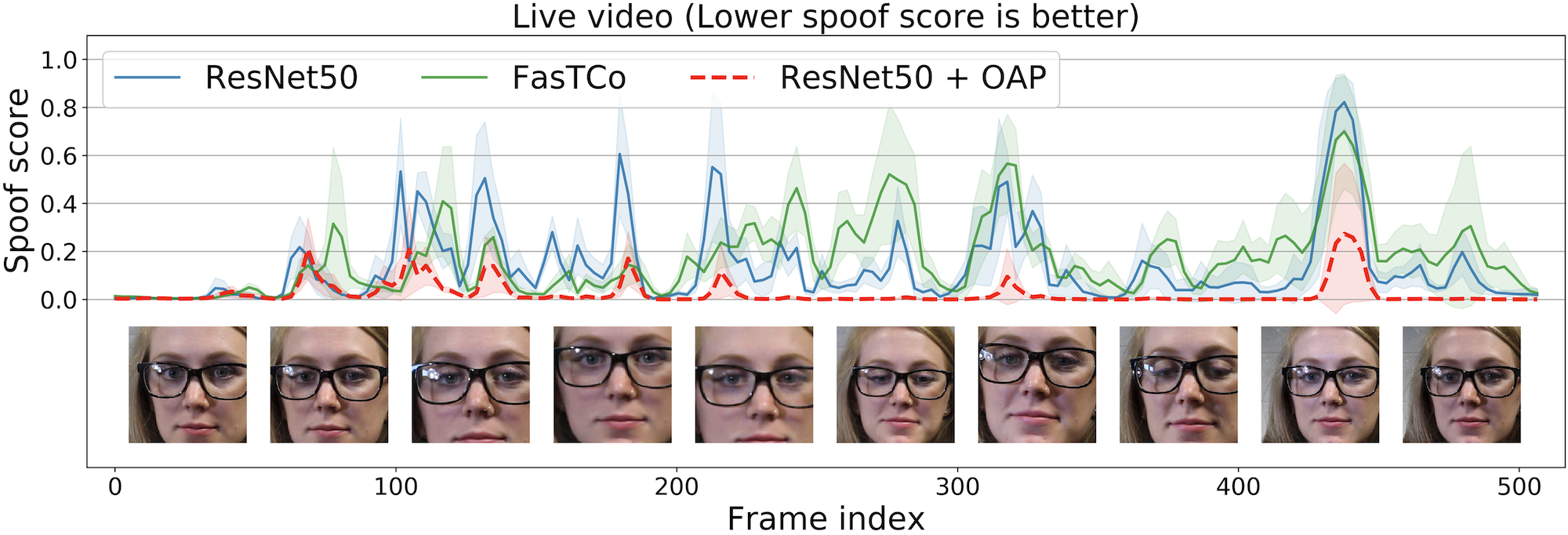}}
   \vspace{2mm}
 \end{minipage}
 
  \begin{minipage}[b]{1.0\linewidth}
   \centering
   \centerline{\includegraphics[width=\textwidth]{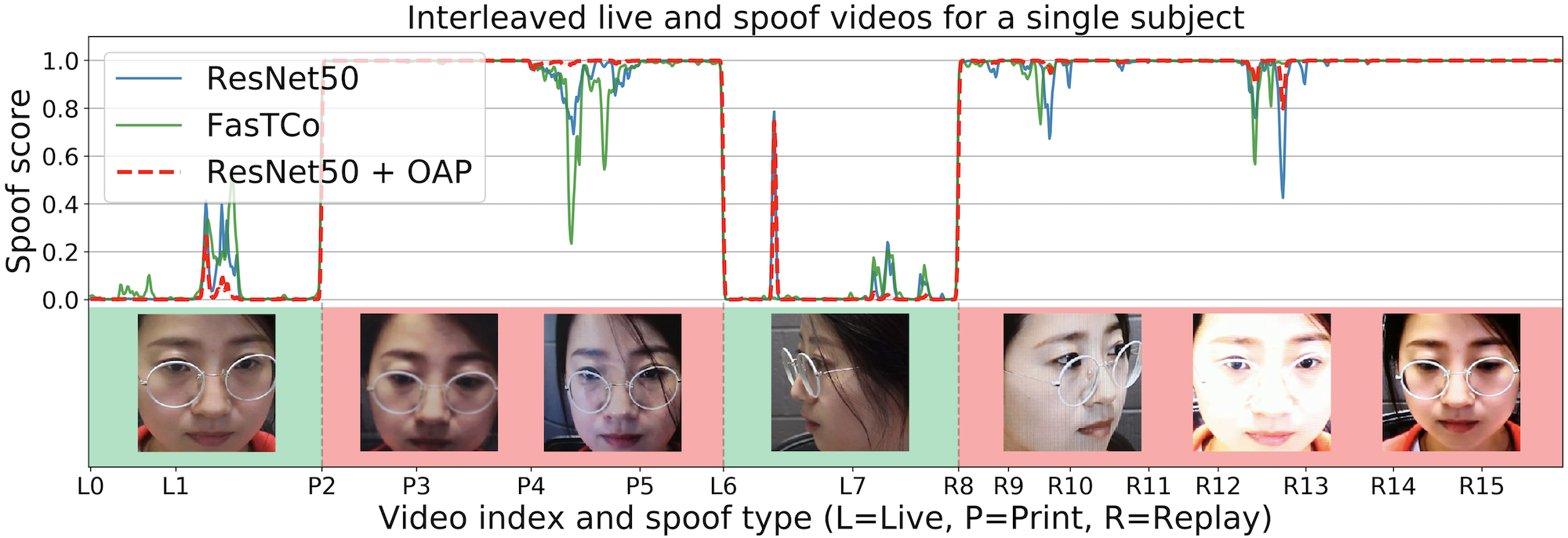}}
 \end{minipage}
  \vspace{-5mm}
 \caption{\textbf{Top:} predictions from ResNet50~\cite{he2016deep}, FasTCo~\cite{fastco} and our proposed OAP in the single video scenario. \textbf{Bottom:} predictions in the continual scenario, where multiple live (in green) and spoof (in red) videos are interleaved and concatenated. In both scenarios the OAP solution adapts over time and outperforms the other methods. 
 }  
 \label{fig:behavior}
 \vspace{-0.5cm}
 \end{figure}

 \begin{figure*}[htb]
\centering
\centerline{\includegraphics[width=\linewidth]{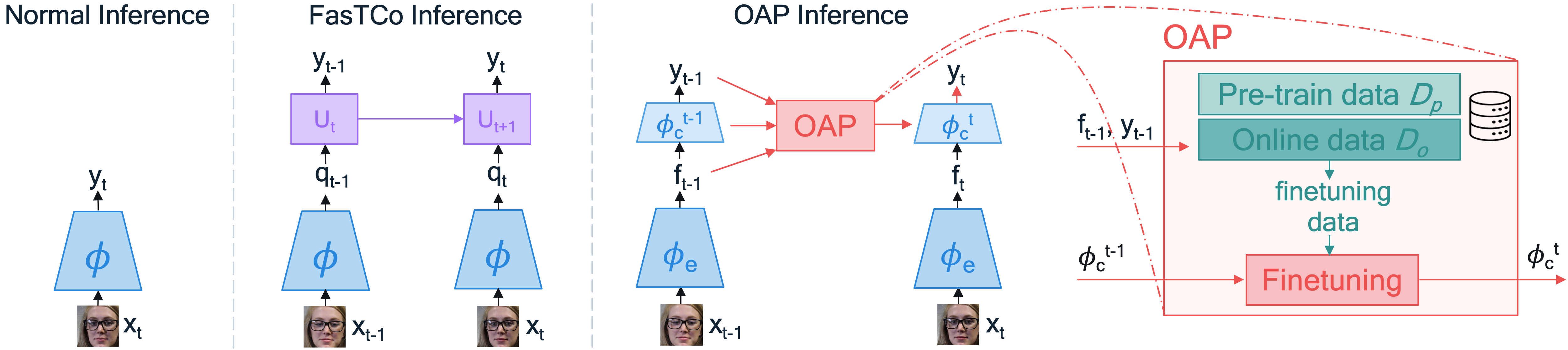}}
\vspace{-1mm}
 \caption{\textbf{Left:} in standard online face anti-spoofing solutions, the prediction for each frame is independent from previous frames. \textbf{Center:} in FasTCo \cite{fastco}, an uncertainty module $U_t$ is used to smooth the logits scores $\mathbf{q}_t$ over time. \textbf{Right:} in OAP (ours), latent features $\mathbf{f}_t$ and predictions $y_t$ are stored and used to adapt parts of the model $\phi_c^t$ over time.}
 \vspace{-6mm}
 \label{fig:method}

 \end{figure*}

While large face anti-spoofing datasets~\cite{liu2018learning,zhang2020celeba,Liu_2021_WACV} have been collected in recent years to enable the development of deep learning solutions, it is infeasible to capture all the variations in the data that might appear at test time. Distribution shift between training and test data naturally occurs due to the presence of new users, sensors, and environmental conditions which were not captured in the training data.
To address the distribution shift due to novel users and sensors, one could use live enrollment data from a user to personalize the face anti-spoofing model to that particular user. For example, in ~\cite{almeida2020detecting, fatemifar2019combining, fatemifar2021client}, the authors use statistics from the enrollment data to calibrate the classification threshold or person-specific coefficients in ensembles. Other recent work ~\cite{wacv22} proposes to obtain personalized predictions by directly conditioning the anti-spoofing model using enrollment images as examples of live data.
However, since the enrollment data is fixed, this type of personalization does not allow continuous adaptation throughout the device's lifetime.

Different approaches for personalization are instead based on domain adaptation and generalization approaches. For example, Shao et al.~\cite{shao2019multi} propose adversarial learning of domain-invariant representations, while Yang et al.~\cite{yang2015person} suggest training subject-specific classifiers with synthesized spoofs for each user. Other works ~\cite{tifs18,aaai21,chen2021generalizable} consider defining specific training objectives to minimize the distribution shift. These personalized methods are also static, as the trained model does not adapt during test time.

Finally, some recent works investigate the adaptation of anti-spoofing models at test time. Quan et al.~\cite{tip21} propose temporal smoothing of predictions and a progressive pseudo-labeling approach where the thresholds are varied over time. Lv et al.~\cite{icassp21} use predictions of model ensembles from different test epochs. However, both of these methods~\cite{tip21,icassp21} assume that all test data is received at once and that the model can be trained using this data before finally making its prediction. This is different from the real-world setting where test data arrives in an online streaming fashion and the system requires low-latency prediction for every incoming frame.
FasTCo~\cite{fastco} proposes an uncertainty-based smoothing method to improve the consistency of predictions over time. While this solution can efficiently run in the online scenario, it does not allow for model adaptation at test time to handle the distribution shift. 

To address the drawbacks found in existing approaches, we propose \textit{OAP} (\textit{Online Adaptive Personalization}): a method to efficiently adapt the anti-spoofing model to specific users and conditions observed at test time. In particular, we develop a solution to address multiple challenges in the online adaptation, like fine-tuning a model with unlabeled data, preventing catastrophic forgetting, and continuously adapting to evolving scenarios. Our solution performs online inference with low latency and minimal compute and memory requirements. In Fig.~\ref{fig:behavior}, we show how the proposed OAP method produces more reliable predictions over time compared to ResNet50 baseline and FasTCo in both single video and continual scenarios.

To summarize, our contributions are as follows: (a) We develop a lightweight personalization method for online face anti-spoofing which runs with negligible compute overhead. Our solution is compatible with most other solutions for face ASP, as it only redefines the behavior of the model at inference time, and does not require storing sensitive personal images on-device; (b) We show consistent improvements with respect to recent anti-spoofing solutions in the online Face ASP scenario for all SiW protocols; (c) We evaluate our approach in a realistic continual interaction setting and observe how our method can handle context switches between live user accesses and spoof attacks without catastrophic forgetting.


\vspace{-2mm}
\section{Proposed Approach}
\vspace{-2mm}
 
\label{sec:approach}


Differently from anti-spoofing via offline video classification, in the online face anti-spoofing scenario \cite{fastco} the model is required to output a low-latency prediction for each frame in the incoming video stream. 
The task can be formulated as the sequential classification of the frames $\mathbf{x}_t$ in a video $\mathbf{V} = [\mathbf{x}_1, \mathbf{x}_2, ..., \mathbf{x}_T]$ of length $T$, with the label for all frames $l_{\mathbf{t}} \in (0=\operatorname{live}, 1=\operatorname{spoof})$ being unknown at test time.
Each frame is evaluated only once to predict its spoof probability $y_t = p(l_t = 1 | \mathbf{x}_t)$.
Commonly, the probability is modeled through a neural network $\phi$ as $y_t = \phi(x_t)$ with parameters optimized on a training dataset and fixed at test time. 

\vspace{-2mm}
\subsection{Pre-training}
\vspace{-1mm}

As our solution adapts the model online, it relies on a pre-trained anti-spoofing model as starting point. Different backbones, sources, and training objectives have been proposed in recent literature to optimize anti-spoofing performance. The proposed OAP solution is compatible with most existing models, as it only assumes that latent features can be obtained at some point in the network. In Fig. \ref{fig:method} we show how the proposed methods performs inference online.

\vspace{-2mm}
\subsection{Online Adaptive Personalization} 
\vspace{-1mm}

Without loss of generality, we represent the anti-spoofing model $\phi$ as two consecutive components: the feature extractor $\phi_f$ and the classifier $\phi_c$, so that $\mathbf{f}_t = \phi_f(\mathbf{x}_t)$ and $y_t = \phi_c(\mathbf{f}_t) = \phi_c(\phi_f(\mathbf{x}_t))$. The features $\mathbf{f}_t$ are low-dimensional latent representations of the input image at time step $t$. In our implementation, we define $\phi_f$ as the convolutional backbone and $\phi_c$ as the final dense layers. During pre-training, both $\phi_f$ and $\phi_c$ are updated, while during online evaluation $\phi_f$ remains fixed and only $\phi_c$ is fine-tuned. By choosing $\phi_c$ to be the last few layers in the network, we ensure that the compute time for updating $\phi_c$ is negligible with respect to time to obtain $\mathbf{f}_t = \phi_f(\mathbf{x}_t)$. In our implementation (see details in Sec. \ref{sec:exp-details}), less than 0.1\% of the compute time during inference is used for OAP. In addition, we only need to store the compact latent features $\mathbf{f}_t$ for fine-tuning instead of the raw full-resolution frames $\mathbf{x}_t$ which contain sensitive biometric information. This design choice allows us to run Online Adaptive Personalization with minimal compute and memory overhead without storing sensitive personal images on the device.

\textbf{Fine-tuning with unlabeled data} \quad The first challenge is how to fine-tune the model online using unlabeled data. Pseudo-labeling allows us to use the model predictions as a proxy for the correct ground-truth labels. A general implementation is: $\; \hat{l}_t = 1 \; \text{if} \; y_t > \tau \; \text{else}$ $0$, with $\hat{l}_t$ being the pseudo-label for frame $\mathbf{x}_t$ and the threshold $\tau$ being calibrated during pre-training. We propose, instead, a more expressive formulation using two thresholds:
\vspace{-1mm}
\begin{equation}
\label{eq:1}
\hat{l}_t = 
\begin{cases}
    1, & \text{if } y_t > \tau_{\operatorname{spoof}}\\
    0,  & \text{if } y_t < \tau_{\operatorname{live}}\\
    \operatorname{discard},& \text{otherwise}
\end{cases}
\vspace{-1mm}
\end{equation}
with $\tau_{\operatorname{spoof}} > \tau_{\operatorname{live}}$ being two separate thresholds for the spoof and live classes. This formulation allows us to discard online samples for which the model predictions are uncertain. We investigate the impact of this formulation through ablation studies in Sec.\ref{sec:experiments}.

To further improve the quality of the pseudo-labels, we exploit once more the sequential aspect of the online data. We can reasonably assume that frames appearing in a small temporal window belong to the same class. Accordingly, a simple smoothing through majority with a sliding window is used to improve the pseudo-label consistency for samples currently stored in the online dataset: 
\vspace{-1mm}
\begin{equation}
\label{eq:2}
\hat{l}_t = \; 1 \; \text{if } \; \Big( \frac{1}{W+1} \sum_{r \in [t-\frac{W}{2}; t+\frac{W}{2}]} \hat{l}_r \Big) > 0.5 \quad \text{else} \; 0. 
\vspace{-1mm}
\end{equation}
We select a time window $W = 30$ frames (equivalent to 1 second in SiW dataset), as we want our solution to work in scenarios with quick transitions between authentic accesses and spoofing attacks.

During online inference, the classifier is initialized with the pre-trained layers $\phi_c^0 = \phi_c$, and the online data is gradually added to an online dataset $D_{\operatorname{o}}$ which is initially empty: $D_{\operatorname{o}}^0 = \varnothing$. For each frame $\mathbf{x}_t$ in the video stream, its features $\mathbf{f}_t$ and pseudo-label $\hat{l}_t$ are used to update the current online dataset $D^{t+1}_{\operatorname{o}} = D_{\operatorname{o}}^t \cup \{(\mathbf{f}_t, \hat{l}_t)\}$. Then, $\phi_c^t$ is fine-tuned using samples from $D^{t+1}_{\operatorname{o}}$ to get the updated model $\phi_c^{t+1}$.
We implement the fine-tuning by few iterations of mini-batch gradient descent with Cross-Entropy objective: 
\vspace{-1mm}
\begin{equation}
\label{eq:3}
L_{\operatorname{CE}}(\hat{l}_i,y_i) = \hat{l}_i \cdot \log y_i + (1 - \hat{l}_i) \cdot \log (1 - y_i),
\vspace{-1mm}
\end{equation}
where $y_i$ are the model predictions and $\hat{l}_i$ are the pseudo-labels for sample $i$ in the online batch.
Next, we describe the challenges and solutions to effectively run OAP online with limited unlabeled data.

\textbf{Preventing catastrophic forgetting} \quad In a real scenario we might expect a user to rarely or never witness spoofing attacks. In this example, only live samples would be provided to the OAP module, and as a result, the model might forget the existence of spoofs. To avoid the catastrophic forgetting of one class or spoof types observed during pre-training, we regularize the adaptation by storing a small subset of the pre-training data $D_{\operatorname{p}}$ to balance the online data $D_{\operatorname{o}}$ during fine-tuning. We perform weighted sampling to balance the distribution of live, spoof, online and offline samples. 
Similar to the online samples, only compressed features $\mathbf{f}$ and labels $l$ from the pre-training samples are required for fine-tuning, which allows for minimal overhead in memory requirements. 

\textbf{Adapting to evolving scenarios} \quad  Finally, we notice that not all past data from the user is needed for the online adaptation. In particular, we would like our anti-spoofing model to be optimized not only for the current user (online personalization) but also for the changing environmental and facial conditions like lighting, pose, and expression (online adaptation). To achieve this, we gradually discard old samples from the online dataset as new frames appear in the online stream. The model will thus gradually adapt to the current conditions. In our implementation, we discard online samples older than 4 seconds, where full videos in the SiW dataset are up to $30$ seconds long. Notice that, even if older samples are discarded, their information will partially be retained in the model's weights thanks to previous OAP iterations.

In Algorithm \ref{alg:cap} we summarize the OAP method in pseudo-code.


\begin{algorithm}
\caption{OAP definition and inference over a sample $\mathbf{x}_t$.}\label{alg:cap}

\begin{algorithmic}[1]
\Function{OAP}{$\mathbf{f}, y, \phi_c, D_{\operatorname{o}}, D_{\operatorname{p}}$}
    \State $\hat{l} \gets$ pseudo\_label$(y, \tau_{\operatorname{spoof}}, \tau_{\operatorname{live}})$
    \Comment Eq.\ref{eq:1}
    \If{$\hat{l} \in \{0, 1\}$}
        \State $D_{\operatorname{o}} \gets D_{\operatorname{o}} \cup \{(\mathbf{f}, \hat{l})\}$
    \EndIf
    \State $D_{\operatorname{o}} \gets$ drop\_old\_samples$(D_{\operatorname{o}})$
    \State $D_{\operatorname{o}} \gets$ label\_smoothing$(D_{\operatorname{o}})$
    \Comment Eq.\ref{eq:2}
    \State $\phi_c \gets $ finetune$(\phi_c, D_{\operatorname{p}}, D_{\operatorname{o}})$
    \Comment Eq.\ref{eq:3}
    \State \Return $\phi_c, D_{\operatorname{o}}$
\EndFunction
\State
\State $\mathbf{f}_t \gets \phi_f(\mathbf{x}_t)$, \;\;\; $y_t \gets \phi_c^t(\mathbf{f}_t)$
\Comment inference
\State $\phi_c^{t+1}, D_{\operatorname{o}}^{t+1} \gets $ \Call{OAP}{$\mathbf{f}_t, y_t, \phi_c^t, D_{\operatorname{o}}^t, D_{\operatorname{p}}$}
\Comment OAP
\end{algorithmic}
\end{algorithm}

\vspace{-4mm}
\section{Experimental Results}
\label{sec:experiments}

\setlength{\tabcolsep}{3pt}
\begin{table*}[t]

\centering
\caption{Experimental results for SiW Protocols 1, 2, 3 averaged over 3 seeds. For all metrics, lower is better. Underlined implies the best result per backbone while bold implies the best result overall.} 
\label{tab:main}
\scalebox{0.88}{
\begin{tabular}{l|cccc|cccc|cccc}  
\toprule
& \multicolumn{4}{c|}{Protocol 1} & \multicolumn{4}{c|}{Protocol 2}  & \multicolumn{4}{c}{Protocol 3} \\
\cmidrule(r){2-13}
 & ACER & APCER & BPCER & EER & ACER & APCER & BPCER & EER & ACER & APCER & BPCER & EER \\
\midrule
RN50 \cite{he2016deep}& 1.12 & 0.37 & 1.98 & 0.75 & 0.61 ± 0.51 & 1.10 ± 1.05 & 0.12 ± 0.06 & 0.34 ± 0.25 & 29.0 ± 12.8 & 57.9 ± 25.6 & \bf{\underline{0.08 ± 0.08}} & 13.8 ± 9.2 \\
RN50 w/ OAP &0.73
  &\underline{0.25}  &1.22  &\bf{\underline{0.49}}
  &\bf{\underline{0.35 ± 0.25}}
  &\bf{\underline{0.57 ± 0.55}}
  &0.13 ± 0.12
  &\bf{\underline{0.18 ± 0.10}}
  &22.9 ± 13.5
  &45.5 ± 27.3
  &0.34 ± 0.34
  &\underline{13.6 ± 8.9}
  \\
RN50 w/ OAP-C &\bf{\underline{0.64}}	&0.55	&\bf{\underline{0.74}}	&0.60	&0.48 ± 0.18	&0.91 ± 0.39	&\bf{\underline{0.04 ± 0.03}}	&0.30 ± 0.05	&\underline{22.8 ± 10.6}	&\underline{45.2 ± 21.5}	&0.41 ± 0.41	&17.4 ± 4.7
 \\
\midrule
FeatherNet \cite{zhang2019feathernets} &1.53	&0.42	&2.64	&0.99	&0.57 ± 0.35	&0.91 ± 0.76	&0.23 ± 0.09	&0.36 ± 0.19	&31.1 ± 11.4	&62.1 ± 22.8	&\underline{0.10 ± 0.08}	&14.0 ± 7.1	
 \\
FeatherNet w/ OAP &\underline{1.00}	&\underline{0.29}	&\underline{1.71}	&\underline{0.87}	&\underline{0.42 ± 0.25}	&\underline{0.67 ± 0.58}	&\underline{0.16 ± 0.10}	&\underline{0.24 ± 0.09}	&\underline{24.3 ± 14.2}	&\underline{48.0 ± 28.9}	&{0.59 ± 0.57}	&\underline{13.8 ± 7.7}
  \\
\midrule
CDCN++ \cite{yu2020searching} &\underline{3.53}	&0.38	
&\underline{6.68}	&\underline{2.32}	&0.84 ± 0.42	&1.48 ± 0.82	&0.20 ± 0.07	&0.61 ± 0.29	&40.2 ± 2.8	&80.2 ± 5.6	&\underline{0.12 ± 0.05}	&25.7 ± 4.3
  \\
CDCN++ w/ OAP &3.69	&\bf{\underline{0.09}}	&7.30	&2.93	&\underline{0.45 ± 0.27}	&\underline{0.88 ± 0.53}	&\underline{0.03 ± 0.02}	&\underline{0.24 ± 0.14}	&\underline{28.7 ± 2.2}	&\underline{54.2 ± 4.1}	&3.05 ± 0.21	&\underline{22.9 ± 1.3}
  \\
\midrule
FasTCo-NA \cite{fastco}&1.08	&0.24	&1.93	&0.64	&\underline{0.56 ± 0.52}	&\underline{1.00 ± 1.05}	&\underline{0.12 ± 0.05}	&\underline{0.32 ± 0.27}	&28.7 ± 13.2	&57.3 ± 26.4	&0.09 ± 0.08	&14.0 ± 9.1  \\
FasTCo \cite{fastco}&\underline{1.05}	&0.23	&\underline{1.86}	&\underline{0.62}	&0.57 ± 0.53	&1.03 ± 1.08	&\underline{0.12 ± 0.05}	&\underline{0.32 ± 0.27}	&28.7 ± 13.2	&57.3 ± 26.5	&\bf{\underline{0.08 ± 0.08}}	&13.8 ± 9.0  \\
FasTCo w/ OAP  &4.40	&\underline{0.20}	&8.61	&1.77	&0.85 ± 0.64	&1.46 ± 1.30	&0.24 ± 0.09	&0.51 ± 0.34	&\bf{\underline{21.7 ± 13.0}}	&\bf{\underline{42.5 ± 26.4}}	&0.88 ± 0.48	&\bf{\underline{12.0 ± 6.3}}
 \\
\bottomrule
\end{tabular}}
\vspace{-4mm}

\end{table*}
\vspace{-2mm}
We compare the proposed method against recent anti-spoofing solutions in a single video and continual scenarios, and we conduct ablation studies on the main hyper-parameters, analysing their impact on performance, compute and memory requirements.

\vspace{-2mm}
\subsection{Experimental setup}
\vspace{-1mm}
\label{sec:exp-details}

\textbf{Dataset} \quad We evaluate the proposed method on the SiW~\cite{liu2018learning} dataset, which consists of 4620 live and spoof videos from 165 subjects collected under different poses and illumination conditions. The dataset defines 3 protocols to measure different generalization capabilities. 

\textbf{Implementation details} \quad To evaluate the efficacy of OAP, we apply it on top of the following baselines: (a) ResNet50~\cite{he2016deep} pre-trained on ImageNet \cite{imagenet}; (b) FeatherNet~\cite{zhang2019feathernets}, a lightweight architecture for anti-spoofing; (c) CDCN++~\cite{yu2020searching} which predicts facial depth maps; and (d) FasTCo~\cite{fastco}, which uses non-adaptive uncertainty based smoothing for inference. We implement all backbones following the hyper-parameters suggested in the original papers, and we set $\phi_c$ to be a dense classifier with a single 64-neurons hidden layer. We train all models with a batch size of 128 over 30k iterations. We use Adam \cite{kingma2015adam} optimizer with a weight decay of $1\text{e-}3$, an initial learning rate of $1\text{e-}3$ and an exponential decay with $\gamma=0.8$ every 1000 iterations. For online adaptation, we use batches of size 16 and learning rate $1\text{e-}6$. 
We evaluate two variants of the proposed methods: OAP, which we evaluate in the standard online scenario over single videos, and OAP-C, which we evaluate in the continual scenario, where live and spoof videos from the same subject are interleaved to simulate spoofing attacks.

\textbf{Evaluation metrics} \quad We report the following metrics for evaluation: (a) APCER: Attack (spoof) Presentation Classification Error Rate; (b) BPCER: Bonafide (live) Presentation Classification Error Rate; (c) ACER: average of APCER and BPCER errors; (d) EER: error rate at which APCER is equal to BPCER.
As the SiW dataset does not include a development set for calibration, we fix the evaluation threshold at 0.5. We also find that using a subset of the training data for threshold calibration yields inaccurate thresholds. While APCER, BPCER, and ACER depend on this threshold, EER allows measuring the discriminative power of the model independently of the evaluation threshold.
During online inference, the evaluation of each frame happens before it is potentially employed for fine-tuning. 

\vspace{-2mm}
\subsection{Quantitative evaluation}
\vspace{-1mm}

The proposed OAP method can be easily applied on top of existing anti-spoofing backbones. In Table \ref{tab:main}, we evaluate OAP in combination and comparison with different backbones. We first observe how the best results across all protocols are obtained with models augmented through OAP, in particular using the ResNet50 backbone. 

We also notice how the improvements obtained by applying OAP on top of ResNet50 and FeatherNet are more consistent and significant with respect to CDCN++ and FasTCo counterparts. We believe this is because the pre-training objective is identical for the first two methods to the one used for fine-tuning in OAP. This is not the case for CDCN++ and FasTCo, where multiple sources and different loss components are used during pre-training.

Finally, we verify that the adaptive model evaluated in the continual scenario (ResNet50 w/ OAP-C) achieves comparable results with the single video evaluation, albeit the hyper-parameters were tuned only in the single video scenario. This serves as proof that the OAP solution does not overfit the specific video on which it is fine-tuned, but it can instead continuously adapt and improve over multiple live and spoof videos without catastrophic forgetting. Next, we further analyse the model behavior in the two scenarios.

\vspace{-2mm}
\subsection{Model behavior over videos}
\vspace{-1mm}

In Fig. \ref{fig:behavior} we visualize the evolution of model predictions over randomly sampled test videos.
In the top plot, we consider the single video scenario and report predictions from ResNet50 and FasTCo in comparison with ResNet50 w/ OAP (plotting average and standard deviation over 3 runs). We notice how the OAP solution quickly adapts the model to the current subject and outputs more confident and robust predictions over time in comparison to both baselines. 

In the bottom plot, we consider the continual scenario, where live and spoof videos from the same subject are concatenated and interleaved. This allows us to simulate a challenging and realistic online scenario, where a user is interacting with the device and, eventually, spoofing attacks take place with multiple source videos from different spoof types. The FasTCo baseline is evaluated in the single video scenario since its uncertainty-based module requires resetting in-between videos from different spoof types. The OAP solution outperforms the baselines by significantly reducing the prediction error and uncertainty in challenging parts of the videos. More importantly, this study confirms that the OAP method is able to personalize and adapt the model to evolving scenarios without catastrophic forgetting of the live or spoof class. We don't observe any prediction delay in the switch of regime between live and spoof videos.

\vspace{-2mm}
\subsection{Ablation studies and efficiency}
\vspace{-1mm}

In Table \ref{table:ablation} we investigate the optimal choices for the main hyper-parameters defining our solution.

\begin{table}[!ht]
\centering
\footnotesize
\setlength{\tabcolsep}{4.5pt}
\vspace{-5mm}
\caption{Ablation study for the main OAP hyper-parameters on SiW Protocol 1. Here $\nu:$ fine-tuning frequency; $(\tau_{\operatorname{spoof}}, \tau_{\operatorname{live}}):$ pseudo-labeling thresholds; $\alpha:$ online samples probability for weighted sampling; $|D_{\operatorname{p}}|:$ number of pre-training samples available during fine-tuning. Best ACER per parameter is highlighted in bold.}
\scalebox{1.05}{
\begin{tabular}{l|ccccc}  
\toprule
$\nu$     & 1 & 0.5 & 0.2 & 0.05 & 0.01 \\
\cmidrule(r){1-6}
ACER     & \bf{0.73} & 0.83 & 0.88 & 0.92 & 0.96 \\
KFLOPs/frame    & 960 & 480 & 192 & 48 & 9.6 \\
\midrule
\toprule

$(\tau_{\operatorname{spoof}}, \tau_{\operatorname{live}})$    & (.01, .99)  & (.05, .95)  & (.1, .9)  & (.2, .8) & (.5, .5)  \\  
\cmidrule(r){1-6}
ACER     & \bf{0.73} & 0.77 & 0.87 & 0.90 & 1.22 \\
\midrule
\toprule

$\alpha$     & 1 & 0.9 & 0.8 & 0.6 & 0.3 \\
\cmidrule(r){1-6}
ACER    & 35.55 & \bf{0.73} & 0.77 & 0.83 & 0.92 \\
\midrule
\toprule

$|D_{\operatorname{p}}|$     & 100 & 500 & 1000 & 5000 & 10000 \\
\cmidrule(r){1-6}
ACER    & 1.08 & 0.82 & 0.73 & \bf{0.71} & 0.72 \\
Memory (MB)    & 0.8 & 4 & 8 & 40 & 160 \\
\bottomrule
\end{tabular}
}
\label{table:ablation}
 
\end{table}
\vspace{-1mm}
\textbf{Fine-tuning frequency} The fine-tuning frequency $\nu$ regulates how often to adapt the model in comparison to the video frame rate. In the case of SiW, we find that fine-tuning the model every 100 frames ($\nu=0.01$) is enough to improve over the ResNet50 baseline. Running the OAP more frequently allows the model to adapt more quickly to environmental and pose changes throughout the video, further improving the model predictions. Since we only update a small part of the neural network $\phi_c$, the compute cost to run OAP after each frame is negligible with respect to the $5\operatorname{GFLOPs}$ of the feature extractor $\phi_f$, which is required for the prediction. For this reason, we select the highest frequency $\nu=1$ in all our experiments.

\textbf{Pseudo-labeling thresholds} \quad Different choices for the pseudo-labeling thresholds $(\tau_{\operatorname{spoof}}, \tau_{\operatorname{live}})$ allow tuning the trade-off between the amount of data available for adaptation and the pseudo-label quality. We define the thresholds symmetrically around $0.5$ to simplify their formulation. We find the optimal configuration to be $\tau_{\operatorname{spoof}}=0.01, \tau_{\operatorname{live}}=0.99$, which suggests pseudo-label quality is more important than sample quantity, at least for the high frame rate in SiW videos. Notice also how the approach based on a single threshold $\tau_{\operatorname{spoof}} = \tau_{\operatorname{live}}=0.5$, which does not allow for discarding uncertain predictions, results in significantly worse performance. 

\textbf{Pre-training data} \quad To conclude, we consider the hyper-parameters regulating the impact of pre-training data. We define $\alpha$ as the probability to select online data points when randomly sampling batches for fine-tuning. Higher values of $\alpha$ allow for faster adaptation. However, when we completely exclude pre-training data by setting $\alpha=1$, we observe catastrophic forgetting as the model overfits to the only class available in the online video stream. We select a sampling weight of $\alpha=0.9$ in our implementation.

The number of pre-training samples available to regularize the online adaptation depends on the selected dataset size $|D_{\operatorname{p}}|$. We randomly sub-sample the selected number of frames from the videos in the pre-training dataset. We find that providing enough variations in terms of spoof types, subjects, illumination, and pose conditions results in better OAP performance. Selecting $|D_{\operatorname{p}}| \geq 500$ is enough to achieve this, with larger sizes providing gradually diminishing returns. Since the pre-training samples must be stored in memory, there is a trade-off between error rate and memory requirements. Considering that ResNet50 parameters require around $94$MB of memory, we select $|D_{\operatorname{p}}| = 1000$ as it provides a good balance between performance and memory requirements. In comparison, the size of the online dataset never exceeds $|D_{\operatorname{o}}| = 120$ (less than $1$MB) since we discard online samples older than $4$ seconds.

\vspace{-3mm}
\section{Conclusion}
\vspace{-2mm}
\label{sec:conclusion}
In this paper, we proposed a lightweight adaptive method to personalize a pre-trained face anti-spoofing model to videos of a specific user. Our method does not require storing raw original images on the device and supports evaluation in the online anti-spoofing scenario. Empirical results confirm that our method can be applied on top of existing solutions to achieve a drop in error rates in both single video and continual settings. We described our solution in detail and included ablation studies for the main hyper-parameters and efficiency costs to validate our implementation choices. 
\clearpage
\bibliographystyle{IEEEbib}
\bibliography{ref}

\end{document}